\documentclass{article}
\usepackage{amsmath, amssymb, amsthm}
\usepackage{geometry}
\usepackage{hyperref}
\usepackage{mathtools}
\usepackage{authblk}
\usepackage{algorithm}
\usepackage{algpseudocode}

\newcommand{\R}{\mathop{\mathbb{R}}}
\newcommand{\K}{\mathcal{K}}

\newtheorem{claim}{Claim}
\newtheorem{observation}{Observation}
\newtheorem{example}{Example}

\title{The Hessian of tall-skinny networks is easy to invert}
\author{Ali Rahimi}

\begin{document}

\maketitle \begin{abstract} We describe an exact algorithm to solve linear
  systems of the form $Hx=b$ where $H$ is the Hessian of a deep net. The method
  computes Hessian-inverse-vector products without storing the Hessian or its
  inverse. It requires time and storage that scale linearly in the number of
  layers.  This is in contrast to the naive approach of first computing the
  Hessian, then solving the linear system, which takes storage and time that are
  respectively quadratic and cubic in the number of layers.  The
  Hessian-inverse-vector product method scales roughly like Pearlmutter's
  algorithm for computing Hessian-vector products.
\end{abstract}

\section{Introduction}

The Hessian of a deep net is the matrix of second-order mixed partial
derivatives of its loss with respect to its parameters. Decades ago, when deep
nets had only hundreds or thousands of parameters, the Hessian matrix could be
inverted to implement optimizers that converged much faster than gradient
descent \cite{Watrous1987, Barnard1992}.  But for large modern deep nets, relying on the
Hessian has become impractical: The Hessian of a model with a billion
parameters would have a quintillion entries, which is far larger than can be
stored, multiplied, or inverted even in the largest data centers. A
common workaround is to approximate the Hessian as a low-rank matrix
\cite{Webb1988, lbfgs-deep-net} or as a diagonal matrix \cite{becker-lecun-89,
  adagrad, adam}. Such approximations make it easier to apply the inverse of the
Hessian to a vector. This article shows how to compute and apply the inverse of
the Hessian exactly without storing the Hessian or its inverse.
The Hessian-inverse-vector product can be computed in time and storage that
scale linearly with the number of layers in the model, and cubically in the
number of parameters and activations in each layer.

Pearlmutter \cite{pearlmutter94} showed how to compute the product of the
Hessian with a fixed vector (the so-called Hessian-vector product) in time and
storage that scale linearly with the number of layers in the network.  This is
much faster than the cubic scaling of the naive algorithm that first computes
the Hessian matrix and then multiplies it by the vector.  His method transforms
the original network into a new network whose gradient is the desired
Hessian-vector product. To compute the Hessian-vector product, one then just
applies backpropagation to the new network.  Given a way to compute the
Hessian-vector product, one can indirectly compute the Hessian-inverse-vector
product via, say Krylov iterations like Conjugate Gradient as proposed by
Pearlmutter and more recently re-investigated \cite{martens-hessian-free, krylov-subspace-descent}. However,
the quality of the result would then depend on the conditioning of the Hessian,
which is notoriously poor for deep nets \cite{ying-behrooz-hessian-spectrum}.
Unfortunately, there seems to exist no variant of Pearlmutter's trick to compute
the Hessian-inverse-vector products directly.

The proposed Hessian-inverse-vector product algorithm takes advantage of a deep
net's layerwise structure.  Regardless of the specific operations in each layer,
the Hessian is a second order matrix polynomial that involves the first order
and second order mixed derivatives of each layer, and the inverse of a block
bi-diagonal operator that represents the backpropagation algorithm.  Multiplying
a vector by this structure computes the Hessian-vector product using exactly the
same operations as Pearlmutter's algorithm (see Appendix \ref{sec:pearlmutter}).
It also leads to a way to compute Hessian-inverse-vector product that does not
require forming or storing the full Hessian. To formally characterize the
storage and running time of the algorithm, assume an oracle that offers the
first order and second order mixed derivatives of each layer.  For an $L$-layer
deep net where each layer has at most $p$ parameters and generates at most $a$
activations, naively storing the Hessian would require $O(L^2p^2)$ memory, and
solving a linear system would require $O\left(L^3p^3\right)$ operations in
addition to the oracle queries. In contrast, we will show how to perform these
operations using only $O(L \max(a, p)^2)$ storage and $O(L \max(a, p)^3)$
computation in addition to the oracle queries. The dependence on the number of
activations and parameters in each layer remains cubic, but the dependence on
the number of layers is only linear. This makes operating on the Hessian of tall
and skinny networks more efficient than the Hessian of short and fat networks.

Although modern networks are typically short and wide, our method runs faster
than the naive Hessian-inverse-vector product algorithm on tall and skinny
networks, since it ameliorates the naive algorithm's cubic dependence on depth
to a mild linear dependence on depth.  A follow-up article will explore training
tall-skinny networks using the Hessian-inverse-vector product as a
preconditioner.  We expect that a significant training speedup will motivate a
return to deeper network designs.

\section{Overview}

Our objective is to efficiently solve linear systems of the form $H x = b$,
where $H$ is the Hessian of a deep neural network, without forming $H$
explicitly. To do this, we employ the following strategy:
\begin{enumerate}
  \item
        Write down the gradient of the deep net in matrix form, as a block bi-diagonal system of linear
        equations. Solving this system uses back-substitution, which in this case
        coincides exactly with the computations carried out by backpropagation.
  \item
        Differentiate this matrix form to obtain an expression for the Hessian. This expression
        involves a second-order polynomial in the inverse of the aforementioned block bi-diagonal matrix.
  \item
        Refactor the expression wit hthe help of auxiliarry variables. These
        lift the polynomial into a higher-dimensional linear form. After
        pivoting, this linear form becomes block tri-diagonal.
  \item
        Factorize the resulting block-tri-diagonal into a lower block
        bi-diagonal matrix, a block diagonal matrix, and an upper block diagonal
        matrix using the LDU factorization.
  \item Solve the resulting system using forward and backward substitution.
        This is similar to running backpropagation on a modified version of the
        original network that's designed to compute the Hessian-inverse-vector
        product.
  \item Un-pivot the result, and report the un-lifted solution.
\end{enumerate}

Algorithm \ref{alg:hessian_inverse_product} summarizes the steps above.

\begin{algorithm}\label{alg:hessian_inverse_product}
  \caption{Compute the hessian-inverse-vector product by solving $(H+\epsilon I)x=b$.}
  \label{alg:hessian_inverse}
  \begin{algorithmic}[1]
    \Require A vector $b$, damping constant $\epsilon$.
    \Ensure $x = (H+\epsilon I)^{-1} b$
    \State Define the sparse block matrix $\mathcal{K}$ and the augmented system as in Equation \ref{eq:unified-system}.
    \State Pivot into permuted system $\mathcal{K}' x' = b'$ where $\mathcal{K}' = \Pi \mathcal{K} \Pi^\top$ is block-tri-diagonal following Equation (\ref{eq:K-pivoted}).
    \State Solve $\mathcal{K}' x' = b'$ using block-LDU decomposition.
    \State Recover the solution $x$ from the permuted vector $\Pi \left[\begin{smallmatrix}x \\ y \\ z\end{smallmatrix}'\right]$.
  \end{algorithmic}
\end{algorithm}

The algorithm is straightforward to implement with a suitable library of
block matrix operations.\footnote{\url{https://github.com/a-rahimi/hessian}}

\section{Notation}

We write a deep net as a pipeline of functions $\ell = 1, \ldots, L$,
\begin{align}
  z_1    & = f_1(z_0; x_1) \nonumber              \\
         & \ldots \nonumber                       \\
  z_\ell & = f_\ell(z_{\ell-1}; x_\ell) \nonumber \\
         & \ldots \nonumber                       \\
  z_L    & = f_L(z_{L-1}; x_L)
\end{align}

The vectors $x_1,\ldots, x_L$ are the parameters of the pipeline. The vectors $z_1,\ldots, z_L$ are its intermediate
activations, and $z_0$ is the input to the pipeline. The last layer $f_L$ computes the final activations and their
training loss, so the scalar $z_L$ is the loss of the model on the input $z_0$. To make this loss's dependence on $z_0$
and the parameters explicit, we sometimes write it as $z_L(z_0;x)$. This formalization deviates slightly from the
traditional deep net formalism in two ways: First, the training labels are subsumed in $z_0$, and are propagated
through the layers until they're used in the loss. Second, the last layer fuses the loss (which has no parameters) and
the last layer (which does).

We'll assume the first and partial derivatives of each layer with respect to its
parameters and its inputs exist. This poses some complications with ReLU
activations and other non-differentiable operations in modern networks. Notably,
for the Hessian to be symmetric, it must be differentiable everywhere.  We'll
largely ignore this complication and assume that differentiable approximations to these
operations are used.

At the end of each section, we'll count the number of floating point operations required to compute various
expressions. While the derivations do not impose any restrictions on the shape of the layers, for the purpose of this
accounting, we'll assume all but the last $L$ layers have $a$-dimensional activations ($z_\ell \in \R^a$) and
$p$-dimensional parameters ($x_\ell \in \R^p$).

\section{Backpropagation, the matrix way}

We would like to fit the vector of parameters $x = (x_1, \ldots, x_L)$ given a training dataset, which we represent by
a stochastic input $z_0$ to the pipeline. Training the model proceeds by gradient descent steps along the stochastic
gradient $\partial z_L(z_0;x) / \partial x$. The components of this direction can be computed by the chain rule with a
backward recursion:
\begin{align}
  \label{eq:backprop}
  \frac{\partial z_L}{\partial x_\ell} & = \underbrace{\frac{\partial z_L}{\partial z_{\ell}}}_{b_\ell} \underbrace{\frac{\partial z_\ell}{\partial x_\ell}}_{\nabla_x f_\ell}                \\
  \frac{\partial z_L}{\partial z_\ell} & = \underbrace{\frac{\partial z_L}{\partial z_{\ell+1}}}_{b_{\ell+1}} \underbrace{\frac{\partial z_{\ell+1}}{\partial z_\ell}}_{\nabla_z f_{\ell+1}}.
\end{align}
The identification $b_\ell \equiv \frac{\partial z_L}{\partial z_\ell}$, $\nabla_x f_\ell \equiv \frac{\partial z_\ell}{\partial x_\ell}$, and $\nabla_z f_\ell \equiv \frac{\partial z_\ell}{\partial z_{\ell-1}}$ turns this recurrence into
\begin{align}
  \frac{\partial z_L}{\partial x_\ell} & = b_\ell \cdot \nabla_x f_\ell          \\
  b_\ell                               & = b_{\ell+1} \cdot \nabla_z f_{\ell+1},
\end{align}
with the base case $b_L = 1$, a scalar.  These two equations can be written in vector form as
\begin{equation}
  \frac{\partial z_L}{\partial x} =
  \begin{bmatrix}
    \frac{\partial z_L}{\partial x_1} & \cdots & \frac{\partial z_L}{\partial x_L}
  \end{bmatrix}
  =
  \underbrace{
    \begin{bmatrix}
      b_1 & \cdots & b_L
    \end{bmatrix}
  }_{\equiv b}
  \underbrace{
    \begin{bmatrix}
      \nabla_x f_1 &                       \\
                   & \ddots                \\
                   &        & \nabla_x f_L
    \end{bmatrix}
  }_{\equiv D_x},
\end{equation}
and
\begin{equation}
  \begin{bmatrix}
    b_1 & b_2 & b_3 & \cdots & b_{L-1} & b_L
  \end{bmatrix}
  \underbrace{
    \begin{bmatrix}
      I                                                          \\
      -\nabla_z f_2 & I                                          \\
                    & -\nabla_z f_3 & I                          \\
                    &               & \ddots & \ddots            \\
                    &               &        & -\nabla_z f_L & 1 \\
    \end{bmatrix}
  }_{\equiv M}
  =
  \underbrace{
    \begin{bmatrix}
      0 & \cdots & 1
    \end{bmatrix}
  }_{\equiv e_L}.
\end{equation}

Solving for $b$ and substituting back gives
\begin{equation}\label{eq:gradient-matrix-form}
  \frac{\partial z_L}{\partial x} =  e_L M^{-1} D_x.
\end{equation}

The matrix $M$ is block bi-diagonal. Its diagonal entries are identity matrices, and its off-diagonal matrices are the
gradient of the intermediate activations with respect to the layer's parameters. The matrix $D_x$ is block diagonal,
with the block as the derivative of each layer's activations with respect to its inputs. $M$ is invertible because the
spectrum of a triangular matrix can be read off its diagonal, which in this case is all ones.

\section{The Hessian}

To obtain the Hessian, we use similar techniques to compute the gradient of
Equation (\ref{eq:gradient-matrix-form}) with respect to $x$.  The gradient we
computed in Equation (\ref{eq:gradient-matrix-form}) is the unique vector $g$
such that $d z_L \equiv z_L(x+dx) - z_L(dx) \to g(x)^\top dx$ as $dx\to 0$.
Similarly, the Hessian $H$ of $z_L$ with respect to the parameters is the unique
matrix $H(x)$ such that $dg\equiv g(x+dx)- g(x)\to H(x)\; dx$
as $dx \to 0$. Appendix \ref{sec:hessian} shows that the Hessian has the following form:

\begin{claim}
  The Hessian of the loss $z_L$ with respect to the vector of parameters $x$ is given by
  \begin{equation}
    \label{eq:hessian}
    H = D_DD_{xx}  + D_DD_{zx} PM^{-1} D_x + D_x^\top M^{-\top}P^\top D_M D_{xz}+D_x^\top M^{-\top}P^\top D_M D_{zz}P M^{-1}D_x,
  \end{equation}
  with the following matrices:
  \begin{align*}
    D_D    & \equiv
    \begin{bmatrix}
      \underbrace{I \otimes b_1}_{p \times ap} &        &               \\
                                               & \ddots &               \\
                                               &        & I \otimes b_L
    \end{bmatrix}
    ,
    \qquad
    D_M \equiv
    \begin{bmatrix}
      \underbrace{I \otimes b_1}_{a \times a^2} &        &               \\
                                                & \ddots &               \\
                                                &        & I \otimes b_L
    \end{bmatrix}
    ,               \\                                                                                     \\
    P      & \equiv
    \begin{bmatrix}
      0 \\ I & 0 \\ &\ddots \\ &I&0
    \end{bmatrix}
    ,    \qquad
    M       \equiv
    \begin{bmatrix}
      I                                                                                    \\
      \underbrace{-\nabla_z f_2}_{a \times a} & I                                          \\
                                              & -\nabla_z f_3 & I                          \\
                                              &               & \ddots & \ddots            \\
                                              &               &        & -\nabla_z f_L & 1 \\
    \end{bmatrix}
    ,               \\
    D_x    & \equiv
    \begin{bmatrix}
      \underbrace{\nabla_x f_1}_{a\times p} &                       \\
                                            & \ddots                \\
                                            &        & \nabla_x f_L
    \end{bmatrix}
    , \qquad
    D_{xx} \equiv
    \begin{bmatrix}
      \underbrace{\nabla_{xx} f_1}_{ap\times p} &        &                 \\
                                                & \ddots &                 \\
                                                &        & \nabla_{xx} f_L
    \end{bmatrix}
    , \qquad
    D_{xz} \equiv
    \begin{bmatrix}
      \underbrace{\nabla_{xz} f_1}_{a^2 \times p} &        &                 \\
                                                  & \ddots &                 \\
                                                  &        & \nabla_{xz} f_L
    \end{bmatrix}
    ,               \\
    D_{zx} & \equiv
    \begin{bmatrix}
      \underbrace{\nabla_{zx} f_1}_{ap \times a} &        &                 \\
                                                 & \ddots &                 \\
                                                 &        & \nabla_{zx} f_L
    \end{bmatrix}
    ,
    \qquad
    D_{zz}  \equiv
    \begin{bmatrix}
      \underbrace{\nabla_{zz} f_1}_{a^2\times a} &        &                 \\
                                                 & \ddots &                 \\
                                                 &        & \nabla_{zz} f_L
    \end{bmatrix}
    .
  \end{align*}
\end{claim}

Given a vector $x \in \R^{Lp}$, the formula above allows us to compute $H x$ in
$O\left(Lap^2 + L a^2p +La^3\right)$ operations without forming $H$. This cost
is dominated by multiplying by the $D_{xx}$, $D_{zx}$, and $D_{zz}$ matrices.
Appendix \ref{sec:pearlmutter} shows that these operations are exactly the
operations performed in Pearlmutter's trick to compute the
Hessian-vector product.

To solve systems of the form $H x = b$, one could use Krylov methods to
repeatedly multiply by $H$ without forming $H$ (a possibility Pearlmutter
considered \cite{pearlmutter94}).  This would require applying $H$ some number
of times that depends on the condition number of $H$. However, the next section
shows how to solve systems of the form $H^{-1} x = b$ with only $\max(a,p)$
times more operations than are needed to compute $H x$.

\section{Applying the inverse of the Hessian}

The above shows that the Hessian is a second order matrix polynomial in
$M^{-1}$. While $M$ itself is block-bidiagonal, $M^{-1}$ is dense, so $H$ is
dense. Nevertheless, this polynomial can be lifted into a higher order object
whose inverse is easy to compute:

\[
  H  = D_D D_{xx}  + D_D D_{zx} P M^{-1} D_x + D_x^\top M^{-\top} P^\top D_M D_{xz} + D_x^\top M^{-\top} P^\top D_M D_{zz} P M^{-1} D_x.
\]
In general, $H$ is singular. We wish to solve the system $(H + \epsilon I ) x = g$ for $x$. We can convert this dense system
involving inverses into a larger, sparse system by introducing auxiliary
variables. Define
\begin{equation}
  y \equiv M^{-1} D_x x,
\end{equation}
which implies $M y - D_x x = 0$.  Define a second auxiliary variable
\begin{equation}
  z \equiv M^{-\top} \left( P^\top D_M D_{xz} x + P^\top D_M D_{zz} P y \right),
\end{equation}
which implies $M^\top z - P^\top D_M D_{xz} x - P^\top D_M D_{zz} P y = 0$.
With these substitutions, $(H + \epsilon I) x = g$ becomes
\begin{equation}
  \left(D_D D_{xx} + \epsilon I\right) x + D_D D_{zx} P y + D_x^\top z = g.
\end{equation}
Collecting these three linear equations gives us a unified system

\begin{equation}
  \label{eq:unified-system}
  \underbrace{
    \begin{bmatrix}
      D_D D_{xx} + \epsilon I & D_D D_{zx} P         & D_x^\top \\
      -D_x                    & M                    & 0        \\
      -P^\top D_M D_{xz}      & -P^\top D_M D_{zz} P & M^\top
    \end{bmatrix}
  }_{\K}
  \begin{bmatrix}
    x \\ y \\ z
  \end{bmatrix}
  =
  \begin{bmatrix}
    g \\ 0 \\ 0
  \end{bmatrix}.
\end{equation}

Finding $x = \left(H+\epsilon I\right)^{-1}g$ is equivalent to solving this block-linear system and
reporting the resulting $x$. The benefit of doing this is that
this system can be pivoted into a block-tri-diagonal system, which can be solved more
efficiently than Gaussian elimination on $\K$ or $H$.

The pivoting we'll apply reorders the blocks of the variables $x,y,z$ from
$x_1,\ldots,x_L, y_1,\ldots, y_L, z_1,\ldots, z_L$ to
$x_1,y_1,z_1,\dots,x_L,y_L,z_L$. This pivoting operation acts as a kind of
transpose operation on block matrices (its generalization is called a
``commutation matrix'' in \cite{magnus-neudecker}). When applied to $\K$, it turns
it into a block-tri-diagonal matrix.  To get a better feel for this operation,
denote the $ij$th block of $\K$ by $\K_{ij}$ and the $uv$th sub-block of this
block by $\K_{ij,uv}$.  Similarly, $x$ denote a vector conformant with $\K$,
with $x_j$ denoting the column vector that multiplies each block $\K_{\cdot j}$
and denote the $v$th sub-block of $x_j$ by $x_{jv}$, which multiplies $\K_{\cdot
    j,\cdot v}$.  We say that $x$ is $j$-major because as we traverse the entries of
$x$ from top to bottom, the index $j$ increments more slowly than the $v$ index.

Denote by $\Pi$ the permutation that transposes the
blocks with the sub-blocks.  Applying this permutation matrix $\Pi$ to $x$ reorders
it from $j$-major to $d$-major. Applying it to the rows and
columns of $\K$ results in a block matrix $\K' \equiv \Pi
  \K \Pi^\top$ that satisfies $\K_{ij,uv} = \K'_{uv,ij}$.
Section \ref{sec:pivoting-examples} illustrates this operator with some examples.

\begin{observation}
  The permutation $\Pi$ is involutory, meaning $\Pi^{-1} =\Pi$.
\end{observation}

This implies that  to solve a system $\K x = b$, we can instead solve $\K' x' =
  \Pi b$ for $x'$, then report $x = \Pi x'$.

This permutation reduces the bandwidth of certain block matrices:

\begin{observation} When the blocks of $\K$ are banded with bandwidth $w$, $\K'$
  is block-banded with bandwidth $w$.
\end{observation}

The blocks in $\K$ are either block-diagonal, block-upper-bi-diagonal, or
block-lower-bi-diagonal, so $\K'$ is at most block-tri-diagonal. Denote the
$ij$th block of $\K'$ as $B_{ij}$:
\[
  B_{ij} \equiv \begin{bmatrix}
    K_{11,ij} & K_{12,ij} & K_{13,ij} \\
    K_{21,ij} & K_{22,ij} & K_{23,ij} \\
    K_{31,ij} & K_{32,ij} & K_{33,ij} \\
  \end{bmatrix}.
\]
Then $\K'$ can be written as the block-tri-diagonal matrix
\begin{equation}
  \label{eq:K-pivoted}
  \K' = \begin{bmatrix}
    B_{11} & B_{12} & 0      & \cdots &              &              & 0          \\
    B_{21} & B_{22} & B_{23} & \ddots &              &              & \vdots     \\
    0      & B_{32} & B_{33} & B_{34} & \ddots       &              &            \\
    \vdots & \ddots & \ddots & \ddots & \ddots       & \ddots       & \vdots     \\
    0      & \cdots & 0      & 0      & B_{L-1, L-2} & B_{L-1, L-1} & B_{L-1, L} \\
    0      & \cdots & 0      & 0      & 0            & B_{L, L-1}   & B_{L, L}
  \end{bmatrix}.
\end{equation}

That means we can solve the system $\K'x'=b'$ by first decomposing $\K'$ via
block LDU decomposition into the product of a lower-block-bi-diagonal matrix, a
block-diagonal matrix, and an upper-block-bi-diagonal matrix as $LDU x'=b'$,
then solve for $L y = b'$  using back substitution, and finally solve for
$DUx'=y$ using forward substitution. Because these systems are
block-bi-diagonal, solving them looks like forward and backward propagation on a
chain. The running time for these solvers is linear in the depth of the network
and cubic in the dimension of each block $B_{ij}$. Depending on the block, the
dimension of each of these blocks is either in the number of parameters or the
number of activations in the corresponding layer. In all,assuming $a$ and $p$
are the largest activation and parameter count for a layer, the running time
for solving this system is $O\left(L \max(a, p)^3\right)$ operations.

\section{Conclusion}

We have described a method to compute the product of the inverse Hessian of a
deep neural network with a vector.  Compared to the naive method, which stores
the Hessian and requires computation cubic in the depth of the
network, this method does not store the Hessian and requires computation linear
in the depth of the network.

In its final stage, the method relies on a forward and backward substitution to
solve a block tridiagonal system.  This step bears a similarity to Pearlmutter's
method for computing the Hessian-vector product: It can be interpreted as
running backpropagation on a modified version of the original network whose
gradient is the desired Hessian-inverse-vector product.  The downside of this
similarity is that solving large linear systems with LDU factorization is prone
to numerical instability.  An improvement on the proposal is to solve the block
tridiagonal system with an off-the-shelf banded solver.

Our hope is to use this technique as a preconditioner to speed up stochastic
gradient descent.  Since the method's speedup is greatest when the network is
tall and skinny, we hope it might rekindle interest in extremely deep
architectures.

\bibliographystyle{plain}
\bibliography{hessian}

\appendix

\section{The Hessian}\label{sec:hessian}
We use the facts that
$dM^{-1} = -M^{-1} (dM) M^{-1}$ and $b=e_L M^{-1}$ to write
\begin{align}
  dg & = g(x+dx)- g(x)                                             \\
     & =d(e_L M^{-1} D_x) \nonumber                                \\
     & = e_L M^{-1} (dD_x)+ e_L \left(dM^{-1}\right) D_x \nonumber \\
     & = b \cdot dD_x  - e_L M^{-1} (dM) M^{-1} D_x \nonumber      \\
     & = b \cdot dD_x  - b \cdot (dM) M^{-1} D_x
\end{align}

We compute each of these terms separately. As part of this agenda, we will rely on the gradient of tensor-valued
functions $g:\R^d \to \R^{o_1 \times \cdots \times o_k}$. Define this gradient $\nabla_x g(x)\in \R^{(o_1 \cdots o_k)
    \times d}$ as the unique matrix-valued function that satisfies

\begin{equation}
  \mathrm{vec} \left(g(x+dx) - g(x)\right) \to \nabla_x g(x)
  \cdot dx
\end{equation}

as $dx \to 0$. This convention readily implies the Hessian of a vector-valued function: If $g:\R^d \to \R^o$, then
$\nabla_{xx} g(x) \in \R^{o\times d^2}$ is the unique matrix such that $\mathrm{vec} \left(\nabla_x g(x+dx) - \nabla_x
  g(x)\right) \to \nabla_{xx} g(x) \; dx$. This convention also readily accommodates the chain rule. For example, the
gradient of $h(x) \equiv f(g(x))$ for matrix-valued $f$ and $g$ can be written as $\nabla f \nabla g$ as expected. It
also implies partial derivatives like $\nabla_{yz} g$ for $g:\R^{|x|} \to \R^{|g|}$. If $y\in \R^{|y|}$ and $z\in
  \R^{|z|}$ are restrictions of $x \in \R^{|x|}$ to some $|y|$ and $|z|$-dimensional subsets, then $\nabla_z g(x) \in
  \R^{|g| \times |z|}$, and $\nabla_{yz} g(x) = \nabla_y \nabla_z g(x) \in \R^{|g||z| \times |y|}$. See Chapter 6 of
Magnus \& Neudecker \cite{magnus-neudecker} for a deeper treatment of higher order derivatives of vector-valued
functions.

\subsection{The term involving $dD_x$}

The matrix $D_x$ is block-diagonal with its $\ell$th diagonal block containing the matrix $D_\ell \equiv \nabla_x
  f_\ell$. Using the facts that $\mathrm{vec}(ABC) = \left(C^\top \otimes A \right) \mathrm{vec}(B)$, and $(A\otimes
  B)^\top = A^\top \otimes B^\top$, we get
\begin{align}
  b\cdot (dD_x) & =
  \begin{bmatrix}
    b_1 & \cdots & b_L
  \end{bmatrix}
  \begin{bmatrix}
    dD_1 &        &      \\
         & \ddots &      \\
         &        & dD_L
  \end{bmatrix}
  \nonumber         \\
                & =
  \begin{bmatrix}
    b_1 \cdot dD_1 & \cdots & b_L \cdot dD_L
  \end{bmatrix}
  \nonumber         \\
                & =
  \begin{bmatrix}
    \mathrm{vec}\left(dD_1\right)^\top \left(I \otimes b_1^\top\right) &
    \cdots                                                             &
    \mathrm{vec}\left(dD_L\right)^\top \left(I \otimes b_L^\top\right)
  \end{bmatrix}
  \nonumber         \\
                & =
  \begin{bmatrix}
    \mathrm{vec}\left(dD_1\right) \\
    \vdots                        \\
    \mathrm{vec}\left(dD_L\right)
  \end{bmatrix}
  ^\top
  \begin{bmatrix}
    I \otimes b_1^\top &        &                    \\
                       & \ddots &                    \\
                       &        & I \otimes b_L^\top
  \end{bmatrix}
\end{align}

Observe that $\mathrm{vec}\left(dD_\ell\right) = d\,\mathrm{vec} \nabla_x f_\ell(z_{\ell-1}; x_\ell)$ varies with $dx$
through both its arguments $x_\ell$ and $z_{\ell-1}$. Using mixed partials of vector-valued functions described above,
we get
\begin{equation}
  \mathrm{vec} \left(dD_\ell\right) = d\,\mathrm{vec}\left(\nabla_x f_\ell\right) = \left(\nabla_{xx} f_\ell\right)\; dx_\ell  + \left(\nabla_{zx} f_\ell\right)\; dz_{\ell-1}.
\end{equation}

Stacking these equations gives
\begin{align}
  \begin{bmatrix}
    \mathrm{vec}\left(dD_1\right) \\
    \vdots                        \\
    \mathrm{vec}\left(dD_L\right)
  \end{bmatrix}
   & =
  \begin{bmatrix}
    \nabla_{xx} f_1 &        &                 \\
                    & \ddots &                 \\
                    &        & \nabla_{xx} f_L
  \end{bmatrix}
  dx
  +
  \begin{bmatrix}
    \nabla_{zx} f_1 &        &                 \\
                    & \ddots &                 \\
                    &        & \nabla_{zx} f_L
  \end{bmatrix}
  \begin{bmatrix}
    dz_0 \\ \vdots \\ dz_{L-1}
  \end{bmatrix}
  .
\end{align}

Each vector $dz_\ell$ in turn varies with $dx$ via $dz_\ell = (\nabla_x f_\ell) dx_\ell + (\nabla_z f_\ell)
  dz_{\ell-1}$, with the base case $dz_0 = 0$, since the input $z_0$ does not vary with $dx$. Stacking up this recurrence
gives
\begin{align}
  \begin{bmatrix}
    I             &   &                   \\
    -\nabla_z f_2 & I &                   \\
                  &   & \ddots            \\
                  &   & -\nabla_z f_L & 1 \\
  \end{bmatrix}
  \begin{bmatrix}
    dz_1 \\ \vdots \\ dz_{L-1}  \\ dz_L
  \end{bmatrix}
  =
  \begin{bmatrix}
    \nabla_x f_1 &        &              \\
                 & \ddots &              \\
                 &        & \nabla_x f_L
  \end{bmatrix}
  dx.
\end{align}
We can solve for the vector $
  \begin{bmatrix}
    dz_1 \\ \vdots \\ dz_L
  \end{bmatrix}
  = M^{-1} D_x dx$ and use the downshifting matrix
\begin{equation}
  \label{eq:P}
  P \equiv
  \begin{bmatrix}
    0 \\ I & 0 \\ &\ddots \\ &I&0
  \end{bmatrix}
\end{equation}
to plug back the vector $
  \begin{bmatrix}
    dz_0 \\ \vdots \\ dz_{L-1}
  \end{bmatrix}
  =PM^{-1}D_x dx$:
\begin{align}
  \begin{bmatrix}
    \mathrm{vec}\left(dD_1\right) \\
    \vdots                        \\
    \mathrm{vec}\left(dD_L\right)
  \end{bmatrix}
  =
  \left(
  \begin{bmatrix}
    \nabla_{xx} f_1 &        &                 \\
                    & \ddots &                 \\
                    &        & \nabla_{xx} f_L
  \end{bmatrix}
  +
  \begin{bmatrix}
    \nabla_{zx} f_1 \\ &\ddots& \\ &&\nabla_{zx} f_L
  \end{bmatrix}
  P M^{-1} D_x
  \right)dx.
\end{align}

\subsection{The term involving $dM$}

The matrix $dM$ is lower-block-diagonal with $dM_2,\ldots, dM_L$, and $dM_\ell \equiv d \nabla_z f_\ell$. Similar to
the above, we can write
\begin{align}
  b & \cdot (dM) M^{-1} D_x =
  \begin{bmatrix}
    b_1 & \cdots & b_{L-1} & b_L
  \end{bmatrix}
  \begin{bmatrix}
    0                  \\
    -dM_2 & 0          \\
          & \ddots     \\
          & -dM_L  & 0 \\
  \end{bmatrix}
  M^{-1} D_x                  \\
    & = -
  \begin{bmatrix}
    b_2 \cdot dM_2 & \cdots & b_L \cdot dM_L & 0
  \end{bmatrix}
  M^{-1} D_x                  \\
    & = -
  \begin{bmatrix}
    \mathrm{vec} \left(dM_2\right)^\top \left(I \otimes b_2^\top\right) &
    \cdots                                                              &
    \mathrm{vec} \left(dM_L\right)^\top \left(I \otimes b_L^\top\right) &
    0
  \end{bmatrix}
  M^{-1} D_x                  \\
    & =
  -
  \begin{bmatrix}
    \mathrm{vec} \left(dM_1\right) \\
    \vdots                         \\
    \mathrm{vec} \left(dM_L\right)
  \end{bmatrix}
  ^\top
  \begin{bmatrix}
    0                                               \\
    I \otimes b_2^\top & 0                          \\
                       &   & \ddots                 \\
                       &   & I \otimes b_L^\top & 0
  \end{bmatrix}
  M^{-1} D_x                  \\
    & = -
  \begin{bmatrix}
    \mathrm{vec} \left(dM_1\right) \\
    \vdots                         \\
    \mathrm{vec} \left(dM_L\right)
  \end{bmatrix}
  ^\top
  \begin{bmatrix}
    I \otimes b_1^\top \\ &\ddots  \\ && I \otimes b_L^\top
  \end{bmatrix}
  PM^{-1} D_x.
\end{align}

Each matrix $dM_\ell = d \nabla_z f_\ell(z_{\ell-1}; x_\ell)$ varies with $dx$ through both $x_\ell$ and $z_{\ell -1}$
as $d\,\mathrm{vec} \left(M_\ell\right) = \left(\nabla_{xz} f_\ell\right) dx_\ell + \left(\nabla_{zz} f_\ell\right)
  dz_{\ell-1}$. Following the steps of the previous section gives
\begin{align}
  \begin{bmatrix}
    \mathrm{vec}\left(dM_1\right) \\
    \vdots                        \\
    \mathrm{vec}\left(dM_L\right)
  \end{bmatrix}
  =
  \left(
  \begin{bmatrix}
    \nabla_{xz} f_1 &        &                 \\
                    & \ddots &                 \\
                    &        & \nabla_{xz} f_L
  \end{bmatrix}
  +
  \begin{bmatrix}
    \nabla_{zz} f_1 \\ &\ddots& \\ && \nabla_{zz} f_L
  \end{bmatrix}
  P M^{-1} D_x
  \right)dx.
\end{align}

\subsection{Putting it all together}

We have just shown that the Hessian of the deep net has the form
\begin{align}
  H \equiv \frac{\partial^2 z_L}{\partial x^2}
   & = D_D \left(D_{xx} + D_{zx} PM^{-1} D_x\right) + D_x^\top M^{-T}P^\top D_M \left(D_{xz}+D_{zz}P M^{-1}D_x\right)           \\
   & = D_DD_{xx}  + D_DD_{zx} PM^{-1} D_x + D_x^\top M^{-\top}P^\top D_M D_{xz}+D_x^\top M^{-\top}P^\top D_M D_{zz}P M^{-1}D_x.
\end{align}

\section{Equation (\ref{eq:hessian}) is Pearlmutter's Hessian-vector multiplication algorithm}
\label{sec:pearlmutter}

We showed that Equation (\ref{eq:hessian}) makes it possible to compute Hessian-vector product $H v$ in time linear in
$L$. These operations are equivalent to Pearlmutter's \cite{pearlmutter94} algorithm, a framework to compute
Hessian-vector products in networks with arbitrary topologies. This section specializes Pearlmutter's machinery to the
pipeline topology, and shows that the operations it produces coincide exactly with those of Equation
(\ref{eq:hessian}).

Consider a set of vectors $v_1,\ldots,v_L$ that match the dimensions of the parameter vectors $x_1,\ldots,x_L$. Just as
$z_L(x_1,\ldots,x_L)$ denotes the loss under the parameters $w$, we'll consider the perturbed loss $z_L(x_1+\alpha v_1,
  \ldots, x_L+\alpha v_L)$ with a scalar $\alpha$. By the chain rule,
\begin{equation}
  \frac{\partial}{\partial \alpha} z_L(x_1+\alpha_1 v_1, \ldots, x_L+\alpha_L v_L) \bigg|_{\alpha=0} =
  \nabla_x z_L(x_1, \ldots, x_L) \cdot v.
\end{equation}
Applying $\nabla_x$ to both sides gives
\begin{equation}
  \nabla_x \frac{\partial}{\partial \alpha} z_L(x_1+\alpha_1 v_1, \ldots, x_L+\alpha_L v_L) \bigg|_{\alpha=0} =
  \nabla_x^2 z_L(x_1, \ldots, x_L) \cdot v.
\end{equation}
In other words, to compute the Hessian-vector product $\nabla_x^2 z_L \cdot v$,
it suffices to compute the gradient of $\frac{\partial z_L}{\partial \alpha} $
with respect to $x$.  We can do this by applying standard backpropagation
to $\frac{\partial z_L}{\partial \alpha}$. At each stage $\ell$ during its
backward pass, backpropagation produces $\frac{\partial}{\partial x_\ell}
  \frac{\partial z_L}{\partial \alpha} = \nabla_{x_\ell, x} z_L \cdot v$, yielding
a block of rows in $\nabla_x^2 z_L \cdot v$.

To see that this generates the same operations as applying Equation (\ref{eq:hessian}) to $v$, we'll write the backprop
operations from Equation (\ref{eq:backprop}) against $\frac{\partial z_L}{\partial \alpha}$ explicitly. We'll use again
the fact that $z_\ell$ depends on $\alpha$ through both $z_{\ell-1}$ and $x_\ell + \alpha v_\ell$ to massage the
backward recursion for $\frac{\partial z_L}{\partial \alpha}$ into a format that matches Equation
(\ref{eq:backprop}):
\begin{align}
  b'_\ell & \equiv \frac{\partial}{\partial z_\ell} \frac{\partial z_L}{\partial \alpha}
  = \frac{\partial}{\partial \alpha} \frac{\partial z_L}{\partial z_\ell}
  = \frac{\partial}{\partial \alpha} b_\ell
  = \frac{\partial}{\partial \alpha} \left[b_{\ell+1} \cdot \nabla_z f_{\ell+1}\right]                                                                                                                                                 \\
          & = b'_{\ell+1} \cdot \nabla_z f_{\ell+1} + \left[\left(I \otimes b_{\ell+1}\right) \frac{\partial}{\partial \alpha} \text{vec}\left(\nabla_z f_{\ell+1}\right) \right]^\top                                                 \\
          & = b'_{\ell+1} \cdot \nabla_z f_{\ell+1} + \left[\left(I \otimes b_{\ell+1}\right) \left(\nabla_{zz} f_{\ell+1} \cdot \frac{\partial z_\ell}{\partial \alpha} + \nabla_{xz} f_{\ell+1} \cdot v_{\ell+1}\right)\right]^\top.
\end{align}
During the backward pass, from $b_\ell$ and $b'_\ell$, we compute
\begin{align}
  \frac{\partial }{\partial x_\ell} \frac{\partial z_L}{\partial \alpha}
   & =\left(\nabla_{x_\ell, x} z_L\right) \cdot v
  = \frac{\partial}{\partial \alpha} \frac{\partial z_L}{\partial x_\ell}
  = \frac{\partial}{\partial \alpha} \left[\frac{\partial z_L}{\partial z_{\ell}} \frac{\partial z_\ell}{\partial x_\ell}\right]
  = \frac{\partial}{\partial \alpha} \left[b_\ell \cdot \nabla_x f_\ell\right]                                                                                                                             \\
   & = b'_\ell \cdot \nabla_x f_\ell + \left[\left(I\otimes b_\ell\right) \frac{\partial}{\partial \alpha} \text{vec}\left(\nabla_x f_\ell\right)\right]^\top                                              \\
   & = b'_\ell \cdot \nabla_x f_\ell + \left[\left(I\otimes b_\ell\right) \left(\nabla_{zx} f_\ell \cdot \frac{\partial z_{\ell-1}}{\partial \alpha} + \nabla_{xx} f_\ell \cdot v_\ell\right)\right]^\top.
\end{align}
Stacking these backward equations horizontally with $b'\equiv\left[
    \begin{smallmatrix}
      b'_1 & \cdots & b'_L
    \end{smallmatrix}
    \right]$,
$g^\alpha_\ell \equiv \frac{\partial z_\ell}{\partial \alpha}$ and $g^\alpha\equiv\left[
    \begin{smallmatrix}
      g^\alpha_1 \\ \vdots \\ g^\alpha_L
    \end{smallmatrix}
    \right]$, then transposing, gives
\begin{align}
  \label{eq:pearlmutter-2}
  M^\top \left(b'\right)^\top & = P D_M  \left(D_{zz} g^\alpha + D_{xz} v\right) \nonumber                       \\
  \nabla_x^2 z_L \cdot v      & = D_x^\top \left(b'\right)^\top + D_D \left(D_{zx} P g^\alpha + D_{xx} v\right).
\end{align}

$g^\alpha_\ell$ can be computed during the forward pass via
\begin{equation}
  g^\alpha_\ell \equiv \frac{\partial z_\ell}{\partial \alpha}
  = \nabla_z f_\ell \cdot \frac{\partial z_{\ell-1}}{\partial \alpha} + \nabla_x f_\ell \cdot v_\ell
  = \nabla_z f_\ell \cdot g^\alpha_{\ell-1} + \nabla_x f_\ell \cdot v_\ell,
\end{equation}
which when stacked up, gives $M g^\alpha = D_x v$.  Plugging $g^\alpha$ back
into Equation (\ref{eq:pearlmutter-2}) and solving for $b'$ gives
\begin{equation}
  \nabla_x^2 z_L \cdot v = D_x^\top M^{-\top} P D_M  \left(D_{zz} M^{-1} D_x v + D_{xz} v\right) +  D_D \left(D_{zx} P M^{-1} D_x v + D_{xx} v\right).
\end{equation}
This coincides with Equation (\ref{eq:hessian}), showing that the two algorithms are equivalent.

\section{Examples of Pivoting Matrices to Block-Banded Form}
\label{sec:pivoting-examples}

\begin{example}
  Suppose each block $D_{ij}$ of $\K$ is diagonal. Then the full $4\times4$ matrix is
  \[
    \K = \begin{bmatrix}
      D_{11} & D_{12} \\
      D_{21} & D_{22}
    \end{bmatrix} =
    \begin{bmatrix}
      a & 0 & c & 0 \\
      0 & b & 0 & d \\
      e & 0 & g & 0 \\
      0 & f & 0 & h
    \end{bmatrix},
  \]
  where
  \[
    D_{11} = \begin{bmatrix} a & 0 \\ 0 & b \end{bmatrix}, \quad
    D_{12} = \begin{bmatrix} c & 0 \\ 0 & d \end{bmatrix}, \quad
    D_{21} = \begin{bmatrix} e & 0 \\ 0 & f \end{bmatrix}, \quad
    D_{22} = \begin{bmatrix} g & 0 \\ 0 & h \end{bmatrix}.
  \]

  The vector $x = \begin{bmatrix} x_1^1 \\ x_1^2 \\ x_2^1 \\ x_2^2
    \end{bmatrix}$ (in $j$-major order) is reordered by $\Pi$ into
  $\begin{bmatrix} x_1^1 \\ x_2^1 \\ x_1^2 \\ x_2^2 \end{bmatrix}$ ($v$-major order).
  Then
  \[\Pi \K \Pi =
    \begin{bmatrix}
      \hspace{0.3em}
      \begin{smallmatrix}
        \rule{0pt}{2.2ex}
        D_{11, 11} & D_{12, 11} \\
        D_{21, 11} & D_{21, 11}
        \rule[-1.2ex]{0pt}{0pt}
      \end{smallmatrix}
      \hspace{0.8em} &
      \hspace{0.3em}
      \begin{smallmatrix}
        \rule{0pt}{2.2ex}
        D_{11, 12} & D_{12, 12} \\
        D_{21, 12} & D_{21, 12}
        \rule[-1.2ex]{0pt}{0pt}
      \end{smallmatrix}
      \hspace{0.3em}   \\
      \hspace{0.3em}
      \begin{smallmatrix}
        \rule{0pt}{2.2ex}
        D_{11, 21} & D_{12, 21} \\
        D_{21, 21} & D_{21, 21}
        \rule[-1.2ex]{0pt}{0pt}
      \end{smallmatrix}
      \hspace{0.8em} &
      \hspace{0.3em}
      \begin{smallmatrix}
        \rule{0pt}{2.2ex}
        D_{11, 22} & D_{12, 22} \\
        D_{21, 22} & D_{21, 22}
        \rule[-1.2ex]{0pt}{0pt}
      \end{smallmatrix}
      \hspace{0.3em}
    \end{bmatrix}
    =
    \begin{bmatrix}
      a & c & 0 & 0 \\
      e & g & 0 & 0 \\
      0 & 0 & b & d \\
      0 & 0 & f & h
    \end{bmatrix}
  \]
  Notice that the resulting matrix has become block-diagonal, whereas $\K$ had a bandwidth of 2.
\end{example}

\begin{example}
  Now endow each block $D_{ij}$ with an upper off-diagonal entry:
  \[
    D_{11} = \begin{bmatrix} a & \alpha \\ 0 & b \end{bmatrix}, \quad
    D_{12} = \begin{bmatrix} c & \beta \\ 0 & d \end{bmatrix}, \quad
    D_{21} = \begin{bmatrix} e & \delta \\ 0 & f \end{bmatrix}, \quad
    D_{22} = \begin{bmatrix} g & \gamma \\ 0 & h \end{bmatrix}.
  \]
  The full $4\times4$ matrix is
  \[
    \K = \begin{bmatrix}
      D_{11} & D_{12} \\
      D_{21} & D_{22}
    \end{bmatrix} =
    \begin{bmatrix}
      a & \alpha & c & \beta  \\
      0 & b      & 0 & d      \\
      e & \delta & g & \gamma \\
      0 & f      & 0 & h
    \end{bmatrix}.
  \]

  After applying the same permutation $\Pi$ as before, we obtain
  \[
    \Pi \K \Pi^\top =
    \begin{bmatrix}
      a & c & \alpha & \beta  \\
      e & g & \delta & \gamma \\
      0 & 0 & b      & d      \\
      0 & 0 & f      & h
    \end{bmatrix}.
  \]
  The off-diagonal entries in each block $D_{ij}$ are collected in
  the upper-right block of the permuted matrix. The permuted matrix
  is block-banded with bandwidth 2, with a dense upper-right block and a sparse
  lower-left block.
\end{example}

\end{document}